\documentclass[letterpaper, 10 pt, conference]{ieeeconf}  

\IEEEoverridecommandlockouts                              

\usepackage{times}
\usepackage{amssymb}
\usepackage{amsmath}
\usepackage{amsthm}
\usepackage{graphics}
\usepackage{epsfig}
\usepackage{multirow}
\usepackage{nicefrac}
\usepackage{cases}          
\usepackage{subcaption}
\usepackage{mathtools}
\usepackage{xcolor}
\usepackage{algorithm}
\usepackage{algorithmicx}
\usepackage{algpseudocode}
\usepackage{gensymb}
\usepackage{balance}

\newcommand{\tn}[1]{\textnormal{#1}}

\newcommand{\mat}[0]{\begin{bmatrix}}
\newcommand{\mate}[0]{\end{bmatrix}}

\newcommand{\vf}{\mathbf{f}}

\newcommand{\vn}{\mathbf{n}}
\newcommand{\vp}{\mathbf{p}}

\newcommand{\vs}{\mathbf{s}}

\newcommand{\vu}{\mathbf{u}}
\newcommand{\vv}{\mathbf{v}}

\newcommand{\vx}{\mathbf{x}}

\newcommand{\cN}{\mathcal{N}}

\newcommand{\cU}{\mathcal{U}}

\newcommand{\cX}{\mathcal{X}}

\newcommand{\hx}{\hat{\mathbf{x}}}
\newcommand{\hp}{\hat{\mathbf{p}}}
\newcommand{\hv}{\hat{\mathbf{v}}}
\newcommand{\hs}{\hat{\mathbf{s}}}
\newcommand{\pr}{\textnormal{Pr}}

\newcommand\norm[1]{\left\|#1\right\|}              
\newcommand{\diag}{\mathop{\mathrm{diag}}}

\newcommand{\half}{\frac{1}{2}}

\makeatletter
\let\NAT@parse\undefined
\makeatother
\usepackage{hyperref}
\hypersetup{hidelinks}
\title{\LARGE \bf
Robust Vision-based Obstacle Avoidance for Micro Aerial Vehicles \\in Dynamic Environments}
\author{Jiahao Lin$^*$, Hai Zhu$^*$ and Javier Alonso-Mora
\thanks{$^{*}$The authors contributed equally. }
\thanks{This work is supported by the Netherlands Organisation for Scientific Research (NWO), domain Applied Sciences, the Amsterdam Institute for Advanced Metropolitan Solutions and 
ONRG-NICOP-grant N62909-19-1-2027. 
The authors are with the Department of Cognitive Robotics, Delft University of Technology, 2628 CD, Delft, The Netherlands {\tt\small $\{$h.zhu; j.alonsomora$\}$@tudelft.nl}}%
}
\begin{document}

\maketitle              

\begin{abstract}
    In this paper, we present an on-board vision-based approach for avoidance of moving obstacles in dynamic environments. Our approach relies on an efficient obstacle detection and tracking algorithm based on depth image pairs, which provides the estimated position, velocity and size of the obstacles. Robust collision avoidance is achieved by formulating a chance-constrained model predictive controller (CC-MPC) to ensure that the collision probability between the micro aerial vehicle (MAV) and each moving obstacle is below a specified threshold. The method takes into account MAV dynamics, state estimation and obstacle sensing uncertainties. The proposed approach is implemented on a quadrotor equipped with a stereo camera and is tested in a variety of environments, showing effective on-line collision avoidance of moving obstacles.
\end{abstract}

\section{Introduction}\label{sec:intro}

Micro Aerial Vehicles (MAVs) are being deployed in a variety of application domains \cite{Chung2018}, such as search and rescue, industrial inspection, and cinematography. In particular, these applications require MAVs to safely navigate and avoid obstacles in the environments. 
Successful autonomous navigation of MAVs using on-board sensors has been demonstrated in static environments \cite{gao2019flying, tordesillas2020faster, falanga2019fast} or in controlled dynamic environments with an overhead motion capture system \cite{Zhu2019, zhu2019distributed}.

The presence of moving obstacles requires a fast and efficient obstacle detection and tracking strategy to perform obstacle avoidance in real time. Moreover, obstacle sensing and MAV state estimation uncertainties should be accounted for to achieve robust collision avoidance.
In this paper, we present an on-board vision-based approach for robust navigation of MAVs in dynamic environments. Our approach builds upon, and extends, a vision-based obstacle detection and tracking algorithm \cite{Oleynikova2015} and a model predictive controller (MPC) \cite{Zhu2019} to generate feasible and probabilistically safe trajectories for the MAV.

\begin{figure}[t]
    \centering
    \includegraphics[width=0.48\textwidth]{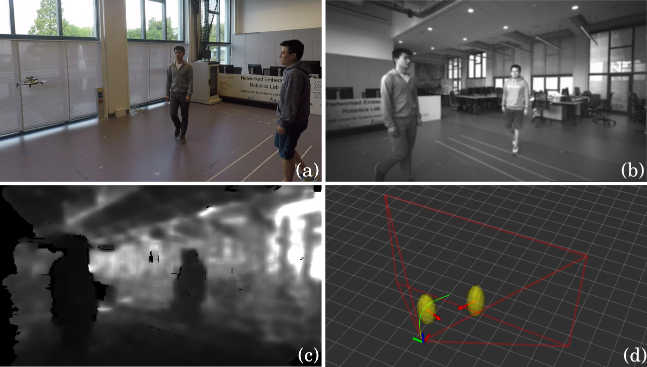}
    \caption{Experimental results of vision-based collision avoidance in dynamic environments with two moving humans. The MAV is equipped with a stereo camera both for visual odometry and obstacle detection. (a) A snapshot of the experiment. (b) On-board grayscale image. (c) The depth image. (d) Visualization of detected obstacles (yellow ellipsoids with red arrows indicating the velocities) and planned collision-free trajectory (green curve).}
    \label{fig:demo_dcsc}
\end{figure}

\subsection{Related Work}\label{sec:relatedWork}
There has been a large amount of work in vision-based autonomous navigation for MAVs in unknown environments \cite{Tang2018}.
Some early works demonstrate autonomous navigation by abstracting simple obstacle representations from camera images and adopting reactive heuristic collision avoidance techniques.  \cite{Ahrens2009} maps obstacles as cylinders by detecting and tracking features of interest from images, then computes a reactive acceleration for collision avoidance. Similarly,  \cite{TomoyukiMori2013} uses a monocular camera to  avoid frontal obstacles by detecting relative size changes of image features and computing an avoidance velocity. However, image processing in those approaches is computationally heavy, which requires an off-board ground computer. In \cite{Biswas2012}, the authors filter 3D point clouds from depth images into planes which are then used to compute the optimal collision avoidance direction with maximal open path length. In \cite{Oleynikova2015}, the authors segment obstacles, modelled as ellipses, from a disparity map and plan a set of waypoints along the edge of obstacles using a heuristic collision checking algorithm. 
While the two approaches are shown to be fast, robot dynamics are not taken into account in the planner. 

Recent works mainly rely on a similar pipeline where an environment map is built from image data and then used to plan collision-free motions for MAVs. In \cite{Fraundorfer2012}, the authors incrementally build a 3D global occupancy map on-board the MAV and use the VHF+ algorithm \cite{Ulrich1998} for collision avoidance. Recent advances have lead to more efficient map representations than the occupancy map \cite{Hornung2013}, including the ESDF map \cite{Oleynikova2017}, the k-d tree \cite{Lopez2017ICRA}, the NanoMap \cite{Florence2018}, the FIESTA map \cite{Han2019IROS}, etc. After building those maps, two main categories of methods are developed to plan collision-free motions. One is to use a library of pre-computed motion primitives \cite{Lopez2017ICRA} or funnels \cite{Majumdar2017} and choose the best one from the library via collision checking. The other is to construct a collision-free flight corridor based a planned path obtained from a discrete planner such as A* and JPS \cite{Liu2017}, followed by trajectory optimization to generate dynamically feasible trajectories for the MAV \cite{Richter2016, Gao2018}. While those approaches have shown successful navigation of a MAV in a variety of environments, a common limitation of them is that they all assume the environments to be static without moving obstacles. Moreover, obstacle sensing uncertainty and MAV state estimation uncertainty are generally neglected.

\subsection{Contribution}
The main contribution of this paper is an integrated system for collision avoidance of moving obstacles in dynamic environments. Obstacles are detected and their position, velocity and size are estimated from depth images and the generated U-depth maps (Section \ref{sec:detection}). Chance-constraints are then formulated to account for the measured MAV state estimation and obstacle sensing uncertainty. These chance constraints are integrated in a model predictive controller to generate dynamically feasible and robust trajectories that keep the probability of collision below a specified threshold (Section \ref{sec:planning}). Finally, we demonstrate the system in real-world experiments to show its effectiveness (Section \ref{sec:results}).

\section{System Overview}\label{sec:system}

Given a goal point, the MAV is required to plan and execute safe collision-free trajectories to navigate towards the goal while avoiding moving obstacles in the environment, based on its sensed stereo depth camera images. 

\begin{figure}[t]
    \centering
    \includegraphics[width=0.47\textwidth]{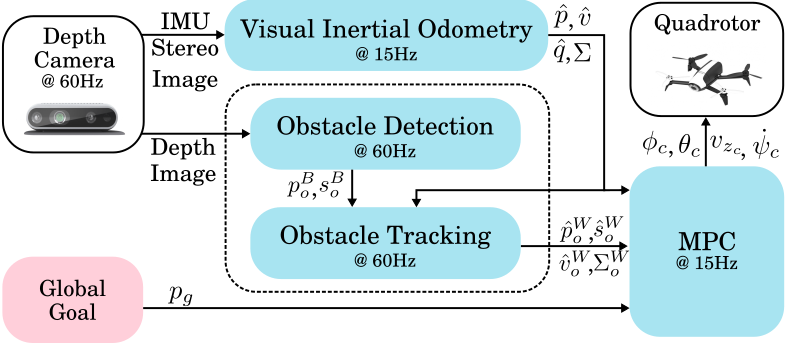}
    \caption{The proposed system for robust vision-based obstacle avoidance in dynamic environments.}
    \label{fig:framework}
\end{figure}

Fig. \ref{fig:framework} illustrates the proposed system for solving the problem, which consists of three main components: MAV state estimation, obstacle sensing and collision-free trajectory planning. In this paper, we focus on the last two components and achieve robust collision avoidance of moving obstacles. For state estimation, we rely on a visual-inertial odometry (VIO) method \cite{Sun2017} to obtain the MAV pose and associated uncertainty. For obstacle sensing, we model obstacles as three-dimensional ellipsoids and adopt an efficient obstacle detection and tracking approach based on depth images, to obtain the obstacle size, position, velocity and associated uncertainty. For collision-free trajectory planning, we formulate a chance-constrained model predictive controller (CC-MPC) \cite{Zhu2019}, taking into account the MAV state estimation and obstacle sensing uncertainty. The CC-MPC ensures that the  collision probability between the MAV and each obstacle is below a specified threshold. 

\section{Obstacle Detection and Tracking}\label{sec:detection}
\begin{figure}[t]
    \centering
    \includegraphics[width=0.47\textwidth]{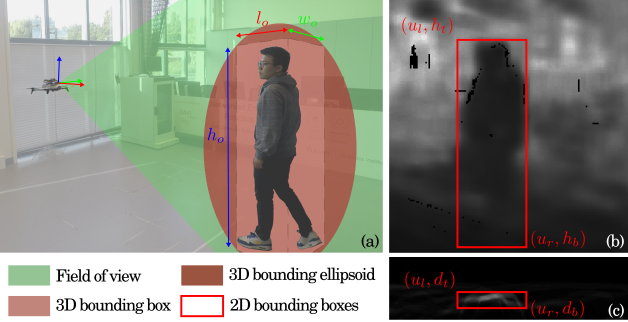}
    \caption{Obstacle detection based on depth image and the U-depth map. (a) Schematic of the camera facing a human obstacle, which is represented as a box with unknown constant size. (b) Obstacle height and vertical ($z$) position detection from the depth image. (c) Obstacle thickness, width and horizontal ($x, y$) position detection from the U-depth map.}
    \label{fig:3d_detection}
\end{figure}

In this section we describe our obstacle detection and tracking algorithm using depth images, as shown in Fig. \ref{fig:3d_detection}. The algorithm is built on \cite{Oleynikova2015} where planar static obstacles are considered. We extend it to three-dimensional scenarios with moving obstacles.

\subsection{Obstacle Detection using Depth Images}\label{subsec:obsDetection}
We model each obstacle as a three-dimensional box with a yaw angle thus facing the camera of the MAV (Fig. \ref{fig:3d_detection}a) and detect its position and size (length/thickness, width and height) based on the camera depth image. The length/thickness and width and horizontal position are firstly derived from the U-depth map (Fig. \ref{fig:3d_detection}c), and then the height and vertical position of the obstacle is derived from the depth image directly (Fig. \ref{fig:3d_detection}b).

\subsubsection{U-depth map}
An U-depth map is computed with the column depth value histograms of the original depth image \cite{Oleynikova2015}. Fig. \ref{fig:3d_detection}b shows an onboard depth image when the MAV is facing a human obstacle in a lab space and Fig. \ref{fig:3d_detection}c is the corresponding generated U-depth map. When an obstacle is in front of the depth camera, the size of the corresponding bin in the U-depth map becomes larger. Based on this property, a bin of the histogram is considered as a point of interest if its value is larger than a threshold $T_{POI}$ defined as 
\begin{align}
    T_{POI}=\frac{fT_{h_{o}}}{d_{bin}},
\end{align}
where $f$ is the focal length, $T_{h_{o}}$ is a predefined threshold for obstacle's height in the space and $d_{bin}$ is the corresponding depth of a bin in the histogram.

Those points of interest are then grouped with other candidate points in their neighborhood so that a bounding box can be extracted from the U-depth map, as shown in Fig. \ref{fig:3d_detection}c.

\subsubsection{Obstacle detection}
We represent obstacles as three-dimensional boxes with unknown, yet constant, sizes. Let $\vp_o^W = (x_o^W, y_o^W, z_o^W)^T$ be the position of the center of a box obstacle, in which the super index $W$ indicates the position expressed in the world frame (while $B$ indicates in the MAV body (camera) frame) and let $ \vs_o^W = (l_o^W, w_o^W, h_o^W)^T$ be the size (length/thickness, width and height) of the box. 

Based on the bounding box found in the U-depth map (see Fig. \ref{fig:3d_detection}c), which is represented by its top-left $(u_{l},d_{t})$ and bottom-right corners $(u_{r},d_{b})$, we can obtain the obstacle's horizontal ($x$ and $y$) position and size (length/thickness and width) in the body frame \cite{Oleynikova2015},
\begin{equation}
    \begin{aligned}
        x_o^B &= d_{b},&y_o^B = \frac{\left(u_{l}+u_{r}\right)d_{b}}{2f},\\ 
        l_o^B &= 2\left(d_{b} - d_{t}\right),&w_o^B = \frac{\left(u_{r}-u_{l}\right)d_{b}}{f}.
    \end{aligned}
\end{equation}

Then we can further find a corresponding bounding box of the obstacle in the original depth image by grouping depth image points whose horizontal index are within $[u_{l},u_{r}]$ and depth values are within $[d_{t},d_{t}+l_o]$. Let $(u_{l},h_{t})$ and $(u_{r},h_{b})$ be the top-left and bottom-right corners of the bounding box. We can derive the obstacle's vertical ($z$) position and height in the body frame,
\begin{equation}
    \begin{aligned}
        z_o^B &= \frac{\left(h_{t}+h_{b}\right)d_{b}}{2f},~~h_o^B &= \frac{\left(h_{t}-h_{b}\right)d_{b}}{f}.
    \end{aligned}
\end{equation}

For a stereo depth camera, the range measurements error generally increases quadratically with the measured depth \cite{Schauwecker2014}. In this paper, we adopt an empirically determined detection uncertainty covariance $\Sigma_o^B$ for the obstacle position and $\Sigma_{o,s}^B$ for the size. 
Then the detected obstacle position, size and corresponding uncertainty covariance are transformed into the world frame by considering the MAV's real-time pose, 
\begin{equation}
    \begin{aligned}
        \vp_o^W &= R_B^W\vp_o^B + \vp^W, ~\Sigma_o^W = R_B^{W^T}\Sigma_o^BR_B^W + \Sigma^W, \\
        \vs_o^W &= R_{B,s}^W\vs_o^B, \quad ~~~ ~~~\Sigma_{o,s}^W = R_{B,s}^{W^T}\Sigma_{o,s}^BR_{B,s}^W,
    \end{aligned}
\end{equation}
where $R_B^W$ is the rotation matrix of the MAV's pose, $R_{B,s}^W = \tn{diag}(\cos{\theta}, \cos{\phi}, \frac{1}{\cos{\theta}\cos{\phi}})$ is the size transfomation matrix to compensate its pitch $\theta$ and roll $\phi$ angles, $\vp^W$ and $\Sigma^W$ denote the position and uncertainty covariance respectively.

\subsection{Obstacle Tracking and Prediction}
To predict obstacle future positions within the prediction horizon, the detected obstacles in sequential frames are firstly associated by evaluating the Gaussian probability density: 
\begin{align}
    pd = p_{G}(\vx_o^{m} ~|~ \hat{\vx}_o^{m|m-1},P_o^{m|m-1}),
\end{align}
where $p_G(\cdot)$ is the probability density function of the multivariate Gaussian distribution,  $\vx_o=(\vp_{o}^{W},\vs_{o}^B)^{T}$ and $P_o = \diag(\Sigma_o^W, \Sigma_{o,s}^B)$ are the obstacle state (position and size) and corresponding uncertainty covariance, the super index $\cdot^m$ indicates the current frame, $\hat{\vx}_o^{m|m-1}$ and $P_o^{m|m-1}$ are the predicted state and covariance based on previous detection by assuming the obstacle is moving at a constant speed. If the probability density $pd$ is larger than a threshold, the two detected objects are determined to be the same moving obstacle whose information is then fed to a Kalman filter.

The Kalman filter estimates the obstacle's position and its velocity and size. Denote by $\hp_o^{k}$, $\hv_o^{k}$ and $\hs_o^{k}$ the estimated obstacle position, velocity and size with uncertainty covariance $\Sigma_o^{k}$, $\Sigma_{o,v}^{k}$ and $\Sigma_{o,s}^{k}$ at time $k$. Here we omit the super index $W$ for simplicity since in the remaining of this paper, all variables are expressed in the world frame.

For collision avoidance of moving obstacles, we predict their future positions and uncertainty covariances using a constant velocity model for obstacle movement. Hence, we have
\begin{equation}\label{eq:obsPredict}
    \begin{aligned}
        \hp_o^{k+1} = \hp_o^k + \hv_o^k\Delta t&, \quad \hv_o^{k+1} = \hv_o^k,  \\
        \Sigma_o^{k+1} = \Sigma_o^k + \Sigma_{o,v}^k\Delta t^2&, \quad \Sigma_o^{k+1} = \Sigma_o^k.
    \end{aligned}
\end{equation}
We assume the size of the obstacle is constant, i.e. $\hs_o^{k+1} = \hs_o^k$, and its uncertainty is not considered in collision avoidance. 

Since polygonal obstacles are ill-posed for online constrained optimization, where smooth shapes are preferred to avoid local minima, we enlarge the detected obstacle box using a bounding ellipsoid with semi-major axes proportional to the box dimensions, i.e.
\begin{align}
    (a_o^k, b_o^k, c_o^k) = \frac{\sqrt{3}}{2}(l_o^k, w_o^k, h_o^k),
\end{align}
and a rotation matrix $R_o^k$ indicating the obstacle orientation (yaw) in the world frame.

\section{Robust Collision Avoidance} \label{sec:planning}

In this section, we present the robust obstacle avoidance method using chance constrained model predictive control (CC-MPC). The method is based on \cite{Zhu2019} which is used for collision avoidance in a controlled environment with an overhead motion capture system. We extend it to an on-board vision based system, by furthering considering the camera's limited filed of view constraints and yaw control of the MAV.

\subsection{Model Predictive Controller}
To formulate the MPC, we first consider the MAV's dynamics model, described by a stochastic nonlinear discrete-time equation,
\begin{align}\label{eq:mavDynModel}
    \vx^{k+1} = \vf(\vx^k, \vu^k) + \omega^k, \quad \vx^0 \sim \cN(\hx^0, \Gamma^0),
\end{align}
where $\vx^k = [\vp^k, \vv^k, \phi^k, \theta^k, \psi^k]^T \in \cX$ denotes the state of the MAV (position, velocity and orienting) and 
$\vu^k \in \cU$ 
the control input at time step $k$. $\cX$ and $\cU$ are the admissible state and control space respectively. The initial state $\vx^0$ is obtained from a state estimator with mean $\hx^0$ and covariance $\Gamma^0$. $\vf$ denotes the nonlinear dynamics. We consider the MAV's motion disturbances as Gaussian process noise $\omega^k \sim \cN(0, W^k)$. 
See \cite{Zhu2019} for details of the dynamics model.

At every time step, for obstacle avoidance, we formulate and solve online a receding horizon constrained optimization problem with $N$ time steps and planning horizon $\tau = N\Delta t$, where $\Delta t$ is the sampling time, as follows, 
\begin{subequations}\label{eq:mpcOpti}
    \begin{align}
    \min\limits_{\hx^{1:N}, \vu^{0:N-1}} \quad          
                            & \sum_{k=0}^{N-1} J^k(\hx^k, \vu^k) + J^N(\hx^N)\label{eq:costFunc} \\ 
    \text{s.t.}	\quad	    & \hx^0 = \hx(t_0), \quad \\
                            & \hx^{k} = \vf(\hx^{k-1}, \vu^{k-1}), \\
                            & \mathbf{G}(\hx^k, \Gamma^k) \leq 0, \label{eq:Constraints}\\ 
                            & \vu^{k-1} \in \cU,~~ \hx^k \in \cX,\\
                            & \forall k\in\{1,\dots,N\},\nonumber
    \end{align}
\end{subequations}
where $J^k$ denotes the cost term at time $k$ and $J^N$ denotes the terminal cost, $\mathbf{G}$ is a function representing the state constraints as described in detail in section \ref{subsec:cons}. $\hx^{k}$ is the mean and $\Gamma^k$ is the uncertainty covariance of the MAV state at time $k$, where the hat $\hat{\cdot}$ denotes the mean of a random variable. We further denote by $\Sigma^k$ the $3\times 3$ covariance matrix of the MAV position $\vp^k$, extracted from $\Gamma^k$.

\subsection{Cost Function}
We now describe the components of the cost function presented in Eq. (\ref{eq:costFunc}). 

\subsubsection{Goal navigation}
Let $\vp^g$ be the given goal position of the MAV. We minimize the displacement between its terminal position in the planning horizon and its goal. To this end, we define the terminal cost term,
\begin{align}
    J^N(\hx^N) = \norm{\hp^N - \vp^g}_{\mathbf{Q}_g},
\end{align}
where $\mathbf{Q}_g$ is a tuning weight coefficient. 

\subsubsection{Control input cost}
The second cost term is to minimize the MAV control inputs, designed as a stage cost,
\begin{align}
    J_u^k(\vu^k) = \norm{\vu^k}_{\mathbf{Q}_u},
\end{align}
where $\mathbf{Q}_u$ is a tuning weight coefficient.

\subsubsection{Collision cost}
To improve flight safety, we also introduce an obstacle potential field cost based on the logistic function. Denote by $d_{o}^k = \norm{\hp^k - \hp_o^k}$ the nominal distance between the MAV and obstacle $o$. Then at time stage $k$, the potential field cost corresponding to obstacle $o$ is
\begin{align}
    J_o^k(\hp^k) = \frac{\tn{Q}_o}{1 + \exp{(\lambda_o(d_o^k - r_o))}},
\end{align} 
where $Q_o$ is a tuning weight coefficient, $\lambda$ is a parameter defining the smoothness of the cost function and $r_o$ is a tuning threshold distance between the MAV and the obstacle where the collision cost is $Q_o/2$. The reason to use a logistic function is to achieve a smooth and bounded collision cost function.

\subsubsection{MAV yaw control}
Since the MAV has a limited field of view, it is generally desirable to make the camera axis, hence the yaw orientation aligned with the direction of motion. Instead of employing a velocity tracking yaw control method as in \cite{Cieslewski2017} which may generate infeasible yaw trajectories, we design a cost function to minimize the deviation between the MAV's yaw and motion direction,
\begin{align}
    J_{\psi}^k(\psi^k) = Q_{\psi}(\psi^k - \bar{\psi}^k)^2,
\end{align}
where $Q_\psi$ is a tuning weight coefficient, $\bar{\psi}^k = \arctan\frac{\hat{v}_y^k}{\hat{v}_x^k}$ indicates the MAV's motion direction angle. To reduce computation time, we compute $\bar{\psi}^k$ based on the MAV's last-loop planned trajectory. 

Finally, the overall stage cost of the formulated MPC is 
\begin{align}
    J^k(\hp^k, \vu^k) = J_u^k(\vu^k) + J_o^k(\hp^k) + J_{\psi}^k(\psi^k).
\end{align}

\subsection{Constraints} \label{subsec:cons}
\subsubsection{Collision chance constraints}
For the obstacle, modelled as an ellipsoid, at position $\vp_o^k$ with semi-principal axes $(a_o^k, b_o^k, c_o^k)$, the MAV at position $\vp^k$ with radius $r$ is considered to be in collision with it if \cite{Zhu2019} 
\begin{align*}
    C_o^k : (\vp^k - \vp_o^k)^T\Omega_o^k(\vp^k - \vp_o^k) \leq 1
\end{align*}
where $\Omega_o^k = R_o^{k,T} \diag(\frac{1}{(a_o^k + r)^2},\frac{1}{(b_o^k + r)^2},\frac{1}{(c_o^k + r)^2}) R_o^k$.

In this paper, we take into account the MAV state estimation uncertainty and obstacle sensing uncertainty. 
Hence, the collision avoidance constraints would be satisfied in a probabilistic manner, which are formulated as chance constraints \cite{Zhu2019, zhu2019mrs}
\begin{align}\label{eq:caChanceCons}
    \pr(C_o^k) \leq \delta, ~\forall k = 1, \dots, N, 
\end{align}
where $\delta$ is the probability threshold for robot-obstacle collision. 

By assuming $\vp^k$ and $\vp_o^k$ are according to Gaussian distributions (obtained from our estimators), i.e. $\vp^k \sim \mathcal{N}(\hp^k, \Sigma^k)$ and $\vp_o^k \sim \mathcal{N}(\hp_o^k, \Sigma_o^k)$, the chance constraint in Eq. (\ref{eq:caChanceCons}) can be transformed into a deterministic constraint with their mean and covariance as follows \cite{Zhu2019}
\begin{equation}\label{eq:detChanceCons}
    \begin{aligned}
        \vn_o^{k^T}\Omega_o^{k^\half}(\hp^k - \hp_o^k) &- 1 \geq \tn{erf}^{-1}(1 - 2\delta) \\
        &\cdot \sqrt{2\vn_o^{k^T}\Omega_o^{k^\half}(\Sigma^k + \Sigma_o^k)\Omega_o^{k^\half}\vn_o^{k}},
    \end{aligned}
\end{equation}
where $\vn_o^{k} = (\hp^k - \hp_o^k)/\norm{\hp^k - \hp_o^k}$, $\tn{erf}(x) = \frac{2}{\sqrt{\pi}}\int_0^xe^{-t^2}dt$ is the standard error function for normal distribution.

\subsubsection{FOV Constraints}
To ensure flight safety, the MAV planned trajectory should be within its current limited field of view (FOV) and limited depth sensing range. Given the MAV's current pose, its three-dimensional FOV with limited depth sensing range can be described by an intersection of five half-spaces,
\begin{align}
    FOV^k := \{ \vp ~|~ \vn_j^k\vp \leq m_j^k \},~j = 1,\dots, 5, 
\end{align}
where $\vn_j^k$ and $m_j^k$ are parameters of the half-spaces. Hence, the FOV constraints are formulated as
\begin{align}
    \vp^k \in FOV^k, ~\forall k = 1, \dots, N.
\end{align}

\subsection{MAV State Uncertainty Propagation}
Evaluating the collision chance constraints in Eq. (\ref{eq:detChanceCons}) requires calculating the MAV state, in particular, position uncertainty covariance at each time step. High-precision uncertainty propagation for nonlinear systems, as in Eq. (\ref{eq:mavDynModel}) could be very computationally intensive \cite{Luo2017}. In this paper, to achieve fast real-time uncertainty propagation, we approximately propagate the MAV state uncertainty using an Extended Kalman Filter (EKF) based update, i.e.
\begin{align}\label{eq:unPropa}
    \Gamma^{k+1} = F^k \Gamma^k F^{k^T} + W^k,
\end{align}
where $W^k$ is the process noise accounting for motion disturbances, $F^k = \frac{\partial\vf^k}{\partial\vx}|_{\hx^{k}, \vu^k}$ is the state transition matrix of the MAV. Then the position uncertainty covariance $\Sigma^k$ can be extracted from $\Gamma^k$. Note that in the above equation, the computation of $F^k$ correlates the robot state and control inputs, which will introduce additional variables into the optimization problem Eq. (\ref{eq:mpcOpti}) and increases the computation time greatly. To this end, we propagate the MAV state uncertainty based on its last-loop trajectory and control inputs before solving this-loop optimization problem.

\section{Results}\label{sec:results}

In this section, we describe our implementation of the proposed approach and evaluate it in real-world experiments. A video showing the flight test results accompanies this paper can be found at \href{https://youtu.be/nZaR-8Z515s}{\color{blue}{https://youtu.be/nZaR-8Z515s}}.

\subsection{Implementation and Hardware}
\begin{figure}[t]
    \centering
    \includegraphics[width=0.32\textwidth]{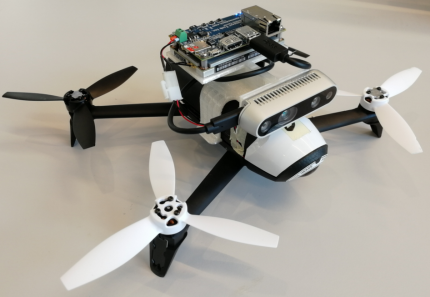}
    \caption{MAV used in the experiments. It is equipped with an NVIDIA Jetson TX2 Module for all on-board computation, an Intel RealSense Depth Camera D435i dually for visual-inertial odometry and depth image sensing.}
    \label{fig:drone}
\end{figure}

Our experimental platform is the Parrot Bebop 2 quadrotor\footnote{\url{https://www.parrot.com/us/drones/parrot-bebop-2}} mounted with an NVIDIA Jetson TX2 Module\footnote{\url{https://developer.nvidia.com/embedded/jetson-tx2}} and an Intel RealSense Depth Camera D435i\footnote{\url{https://www.intelrealsense.com/depth-camera-d435i}}, as shown in Fig. \ref{fig:drone}. The Parrot Bebop 2 allows for executing control commands sent via ROS\footnote{\url{https://bebop-autonomy.readthedocs.io}}. The D435i camera is dually used for visual-inertial odometry and depth image sensing, which has a $87^\circ \times 58^\circ$ FOV and 5 m depth sensing range. The TX2 is used to perform all on-board computation and is connected with the Bebop 2 via WiFi. 

We use a filtering-based stereo visual-inertial odometry algorithm, the S-MSCKF \cite{Sun2017}, for state estimation of the MAV, which runs at 15 Hz. The camera depth images are received at 60 Hz and the obstacle detection and tracking is running at frame rate. We rely on the ACADO toolkit \cite{Houska2011} to generate a fast C solver for our MPC, in which a sampling time of 60 ms is used and the prediction horizon is set to 1.5 s. The radius of the MAV is set as 0.4 m. The two closest detected obstacles are fed to the MPC for collision avoidance. The collision probability threshold is set as $\delta = 0.03$. In order to obtain some quantitative results, in the lab scenarios we use an external motion capture system (OptiTrack) to measure the position of the MAV and moving obstacles, which is only used as ground true data.

\subsection{Obstacle Detection and Tracking Performance}
We first evaluate the obstacle detection and tracking performance for moving obstacles. Fig. \ref{fig:demo_dcsc}(a) shows a lab scene with two walking human obstacles. We put the camera at the origin of the world frame and recorded a dataset of which the two humans were walking around at a speed of approximately 1.2 m/s. Position and velocity of the two human obstacles are obtained using our obstacle detection and tracking algorithm. Table \ref{tab:detection} shows the average position and velocity estimation errors of the two moving obstacles comparing with ground truth measurements. It can be observed that the average position estimation error is around 0.3 m and that of velocity can be up to 0.5 m/s, which indicates the obstacle sensing uncertainty 
should be taken into account when planning robust collision-free trajectories for the MAV. 
In practice, the obstacle's velocity estimation may be very noisy and has a very large uncertainty covariance. In this case we bound the $\Sigma_{o,v}^k$ in Eq. (\ref{eq:obsPredict}) when predicting the obstacle's future positions and corresponding uncertainty covariances.

\begin{table}[t]
    \centering
    \caption{Detection and tracking errors of moving obstacles.}
    \begin{tabular}{ccc}
    \hline
    \multirow{2}{*}{\textbf{\begin{tabular}[c]{@{}c@{}}Moving \\ obstacle\end{tabular}}} & \multicolumn{2}{c}{\textbf{Average estimation error}} \\ \cline{2-3} 
                                                                                         & \textbf{Position (m)}       & \textbf{Velocity (m/s)}       \\ \hline
    \textbf{No. 1}                                                                       & 0.28                        & 0.47                          \\
    \textbf{No. 2}                                                                       & 0.25                        & 0.41                          \\ \hline
    \end{tabular}\label{tab:detection}
\end{table}

\subsection{Obstacle Avoidance in Dynamic Environments}
We tested the system in a variety of flight tests. The results of two typical scenarios are particularly presented here. 

\begin{figure*}[t]
    \centering
    \captionsetup[subfigure]{position=b}
    \begin{subfigure}{0.24\textwidth}
            \includegraphics[width=\textwidth]{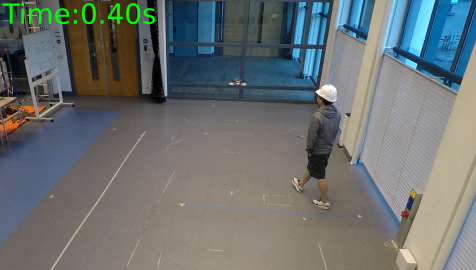}
    \end{subfigure}
    \captionsetup[subfigure]{position=b}
    \begin{subfigure}{0.24\textwidth}
            \includegraphics[width=\textwidth]{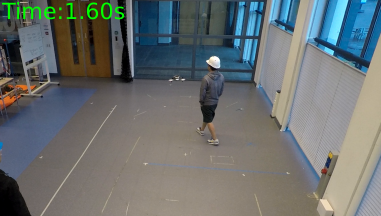}
    \end{subfigure}
    \captionsetup[subfigure]{position=b}
    \begin{subfigure}{0.24\textwidth}
            \includegraphics[width=\textwidth]{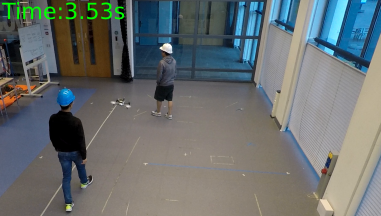}
    \end{subfigure}
    \captionsetup[subfigure]{position=b}
    \begin{subfigure}{0.24\textwidth}
            \includegraphics[width=\textwidth]{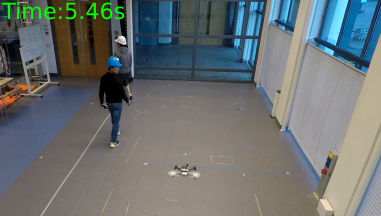}
    \end{subfigure}
    \\ 
    \captionsetup[subfigure]{position=b}
    \begin{subfigure}{0.24\textwidth}
            \includegraphics[width=\textwidth]{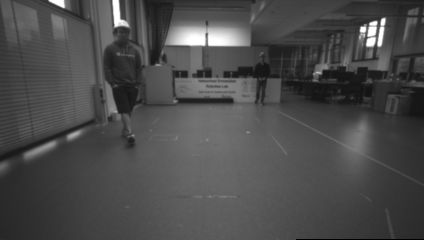}
    \end{subfigure}
    \captionsetup[subfigure]{position=b}
    \begin{subfigure}{0.24\textwidth}
            \includegraphics[width=\textwidth]{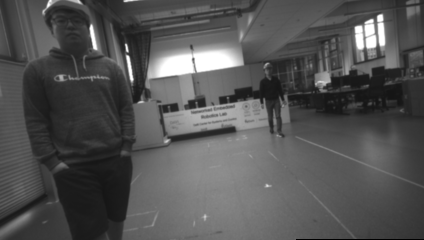}
    \end{subfigure}
    \captionsetup[subfigure]{position=b}
    \begin{subfigure}{0.24\textwidth}
            \includegraphics[width=\textwidth]{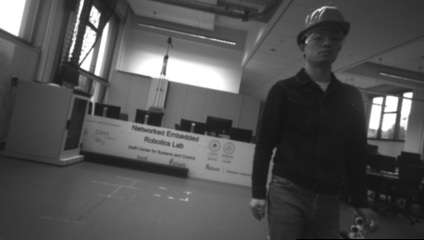}
    \end{subfigure}
    \captionsetup[subfigure]{position=b}
    \begin{subfigure}{0.24\textwidth}
            \includegraphics[width=\textwidth]{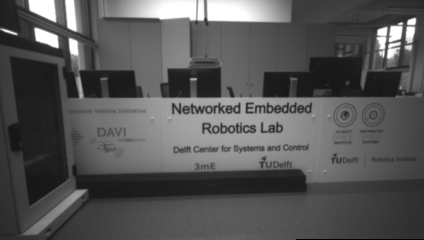}
    \end{subfigure}
    \caption{A sequence of images during the experiment Scenario 1. The MAV is required to fly from a start point to an end goal while avoiding two walking humans. Top: Snapshots of the experiment. Bottom: On-board camera grayscale images.}%
    \label{fig:caDCSC}%
\end{figure*}

\begin{figure*}[t]
    \centering
    \begin{subfigure}{0.3\textwidth}
        \includegraphics[width=\textwidth]{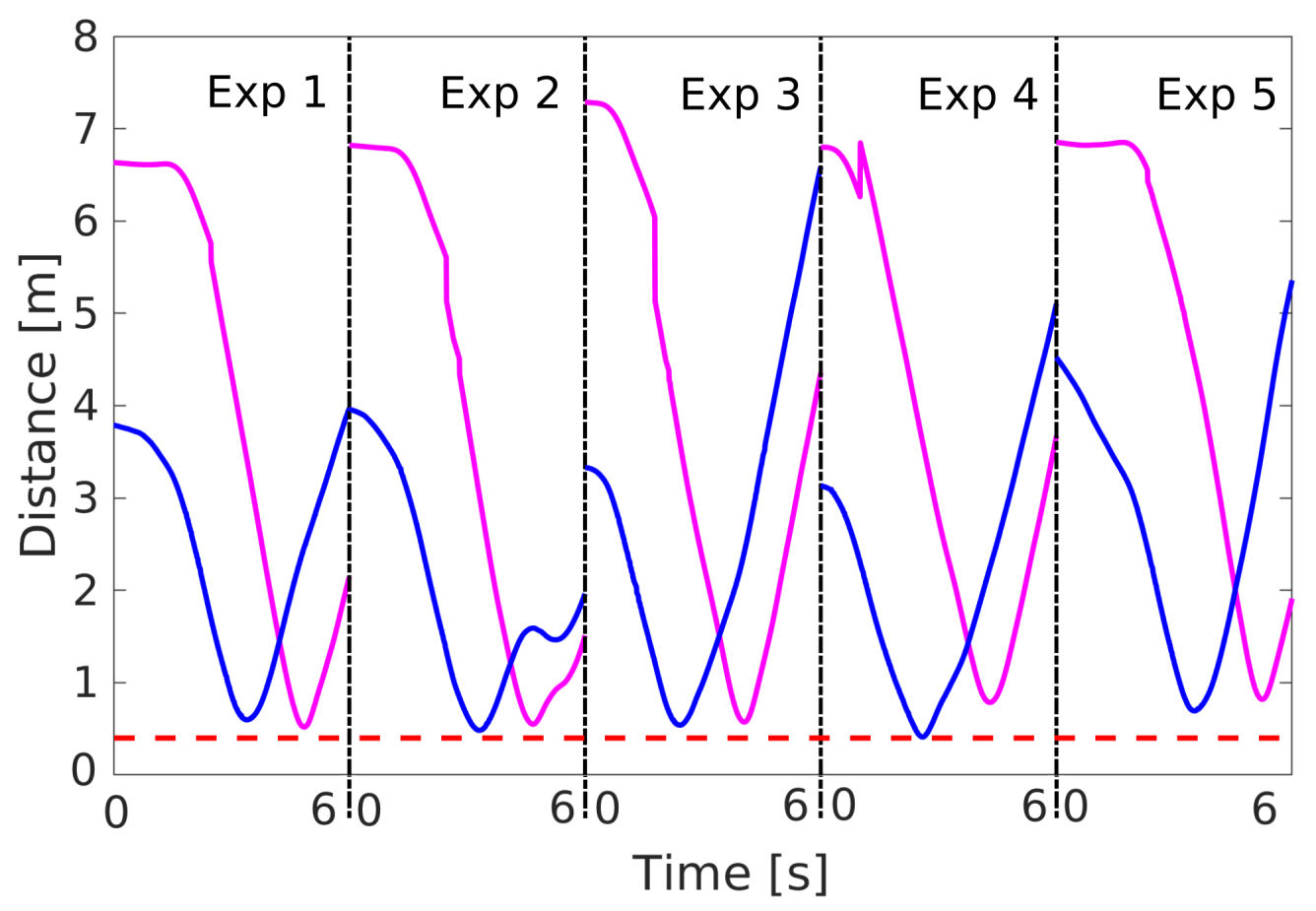}
        \caption{Distance to obstacles over time}
        \label{subfig:distToObs}
    \end{subfigure}
    \begin{subfigure}{0.32\textwidth}
        \includegraphics[width=\textwidth]{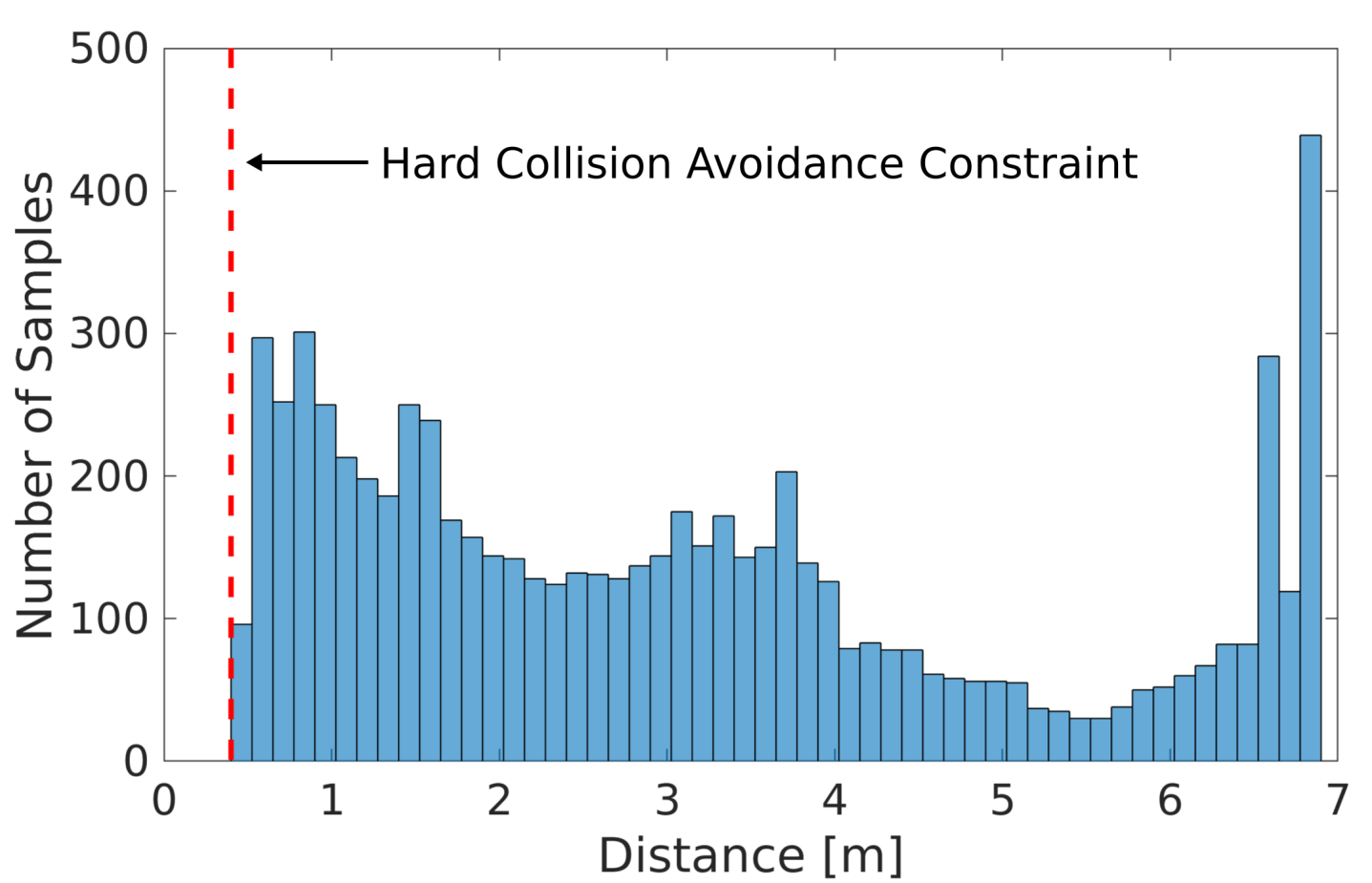}
        \caption{Histogram of distance}
        \label{subfig:distToObsCount}
    \end{subfigure}
    \begin{subfigure}{0.35\textwidth}
        \includegraphics[width=\textwidth]{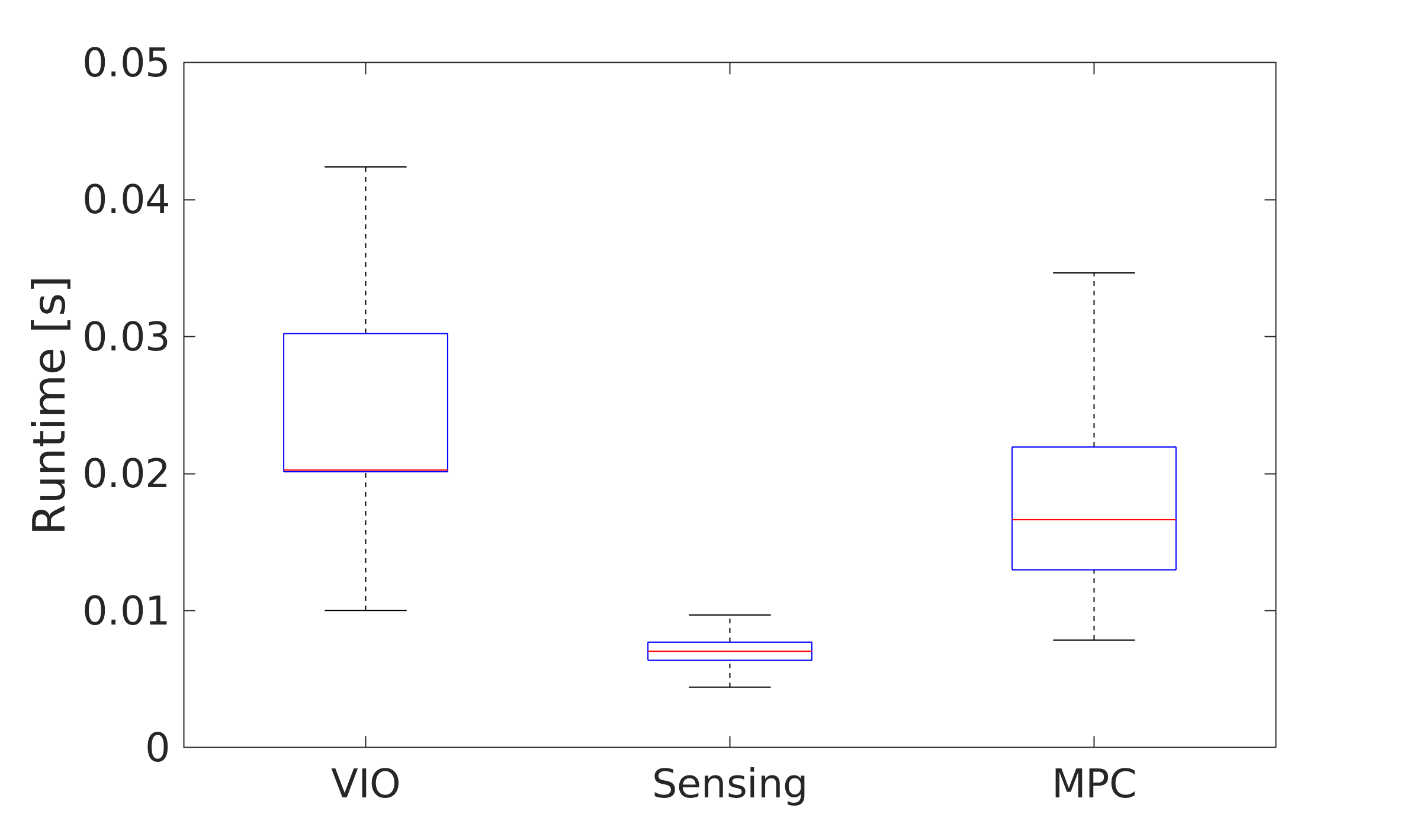}
        \caption{MAV on-board runtimes }
        \label{subfig:runTime}
    \end{subfigure}
    \caption{Quantitative results of the experiment Scenario 1. (a) Distance between the MAV and the two  moving obstacles (magenta and blue) over time during 5 experiments. (b) Histogram of all the distance data. (c) On-board runtimes of the MAV state estimation (VIO), obstacle detection and tracking, and collision-free trajectory optimization (MPC).}\label{fig:resExp1}
\end{figure*}

\begin{figure*}[t]
    \centering
    \captionsetup[subfigure]{position=b}
    \begin{subfigure}{0.32\textwidth}
            \includegraphics[width=\textwidth]{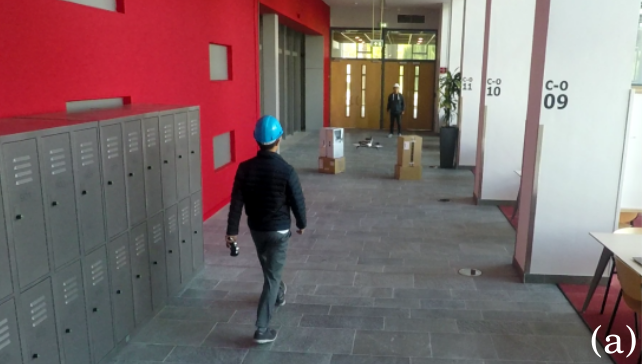}
    \end{subfigure}
    \captionsetup[subfigure]{position=b}
    \begin{subfigure}{0.32\textwidth}
            \includegraphics[width=\textwidth]{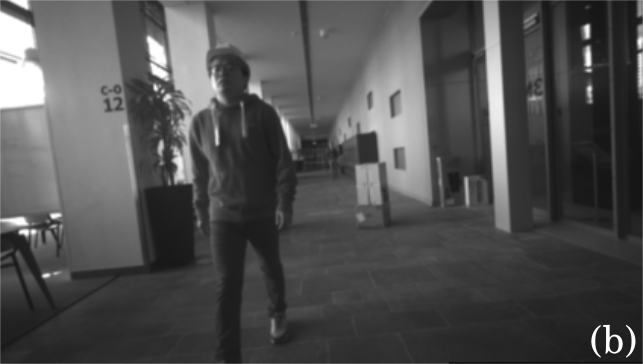}
    \end{subfigure}
    \captionsetup[subfigure]{position=b}
    \begin{subfigure}{0.32\textwidth}
            \includegraphics[width=\textwidth]{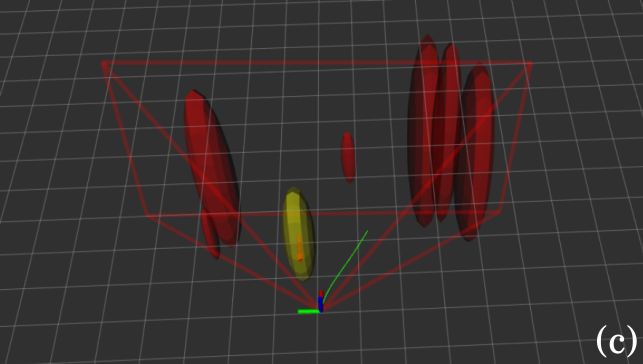}
    \end{subfigure}
    \caption{Results of the experiment Scenario 2. The MAV is flying in a corridor while avoiding static and moving obstacles. (a) A snapshot during the experiment. (b) An on-board grayscale image captured in the experiment and (c) visualization of the corresponding obstacle detection and trajectory planning results.}%
    \label{fig:caCorridor}%
\end{figure*}

\subsubsection{Scenario 1 (Flying in a confined lab space)}
The MAV is required to navigate from a start point to an end point while avoiding two walking humans. Fig. \ref{fig:caDCSC} shows a series of snapshots and the MAV on-board camera grayscale images taken during the experiment. 
In this scenario, we performed the experiment five times. 
Fig. \ref{subfig:distToObs} shows the measured distance between the MAV and the two moving human obstacles over time in the five experiments. 
The distance is computed, based on ground truth measurements, as the closest distance between the MAV's position and the obstacle ellipsoid's surface (with semi-major axis $(0.4, 0.4, 0.9) m$).
In Fig. \ref{subfig:distToObsCount}, we cumulate all the distance data. 
It can be observed that in all instances a minimum safe separation of 0.4 m was achieved and therefore collisions with the humans were avoided. A maximal speed of around 1.6 m/s of the MAV was observed in this experiment.

The boxplots of the on-board runtimes in this scenario is shown in Fig. \ref{subfig:runTime}. For the runtimes of the obstacle detection and tracking, the 75$^{\tn{th}}$ percentile is always below 8 ms, which is fast enough to be run at frame rate (60 Hz). For the runtimes of the MPC framework, the 75$^{\tn{th}}$ percentile is always below 22 ms, indicating the framework can be run efficiently in real time.

\subsubsection{Scenario 2 (Flying in a long corridor)}
The MAV is flying in a long narrow corridor where there are both static and moving obstacles. Fig. \ref{fig:caCorridor} shows a snapshot taken during the experiment. A maximum speed of around 2.4 m/s was achieved by the MAV in the experiment. Detailed results of the experiment can be found in the video accompanying this paper.

\section{Conclusion}\label{sec:conclsuion}

We presented an on-board vision-based obstacle avoidance approach for MAVs navigating in dynamic environments. Flight test results in a variety of environments with moving obstacles demonstrated the effectiveness of the proposed approach. We adopted a fast three-dimensional obstacle detection and tracking algorithm based on depth images which can run at a frame rate of 60 Hz. We took into account the obstacle sensing uncertainty by using a chance constrained model predictive controller (CC-MPC) to generate robust local collision-free trajectories for the MAV. We implemented the approach on a computational power-limited quadrotor platform, where the obstacle detection and tracking has a mean computation time of 8 ms and that of the MPC is 16 ms. In real-world indoor experiments, the MAV is shown to be able to avoid walking human obstacles at a maximum speed of 2.4 m/s.


\bibliographystyle{IEEEtran}
\balance
\bibliography{ref}

\end{document}